\documentclass[runningheads]{llncs}
\usepackage{graphicx}
\usepackage{amsmath}
\usepackage{amssymb}
\usepackage{bbding}
\usepackage{enumitem}
\usepackage{multirow}
\usepackage{booktabs}
\usepackage{bbm}
\usepackage{dsfont}

\usepackage{comment}
\usepackage{subcaption}
\usepackage{caption}

\usepackage{hyperref}

\usepackage{cite}
\usepackage{bm}
\usepackage{color, soul}
\sethlcolor{blue}
\usepackage[table]{xcolor}

\newcommand{\hlr}[1]{{\color{red}{#1}}}
\newcommand{\hlb}[1]{{\color{blue}{#1}}}

\begin{document}
%

\title{CircleFormer: Circular Nuclei Detection in Whole Slide Images with Circle Queries and Attention}

\author{Hengxu Zhang\inst{1,2} \and
Pengpeng Liang\inst{3}\and
Zhiyong Sun\inst{1} \and
Bo Song\inst{1} \and
Erkang Cheng\inst{1} \thanks{Corresponding author. \email{twokang.cheng@gmail.com}}
}

\institute{Institute of Intelligent Machines, HFIPS, Chinese Academy of Sciences, China \and Institutes of Physical Science and Information Technology, Anhui University, China \and 
School of Computer and Artificial Intelligence, Zhengzhou University, China
}

\maketitle              
\begin{abstract}
Both CNN-based and Transformer-based object detection with bounding box representation  have been extensively studied in computer vision and medical image analysis, but circular object detection in medical images is still underexplored. Inspired by the recent  anchor free CNN-based circular object detection method (CircleNet) for ball-shape glomeruli detection in renal pathology, in this paper, we present CircleFormer, a Transformer-based circular medical object detection with dynamic anchor circles. Specifically, queries with circle representation in Transformer decoder iteratively refine the circular object detection results, and  a circle cross attention module is introduced to compute the similarity between circular queries and image features.  A generalized circle IoU (gCIoU) is proposed to serve as a new regression loss of circular object detection as well. Moreover, our approach is easy to generalize to the segmentation task by adding a simple segmentation branch to CircleFormer. We evaluate our method in circular nuclei detection and segmentation on the public MoNuSeg dataset, and the experimental results show that our method achieves promising performance compared with the state-of-the-art approaches.  The effectiveness of each component is validated via ablation studies as well. Our code is released at: \url{https://github.com/zhanghx-iim-ahu/CircleFormer}.

\end{abstract}

\section{Introduction}
  \begin{figure}[h]
 	\begin{center}
		\begin{minipage}[b]{0.45\linewidth}
			\setcounter{figure}{0}
			\begin{minipage}[b]{1\linewidth}
			\begin{center}
        		\includegraphics[width=\linewidth]{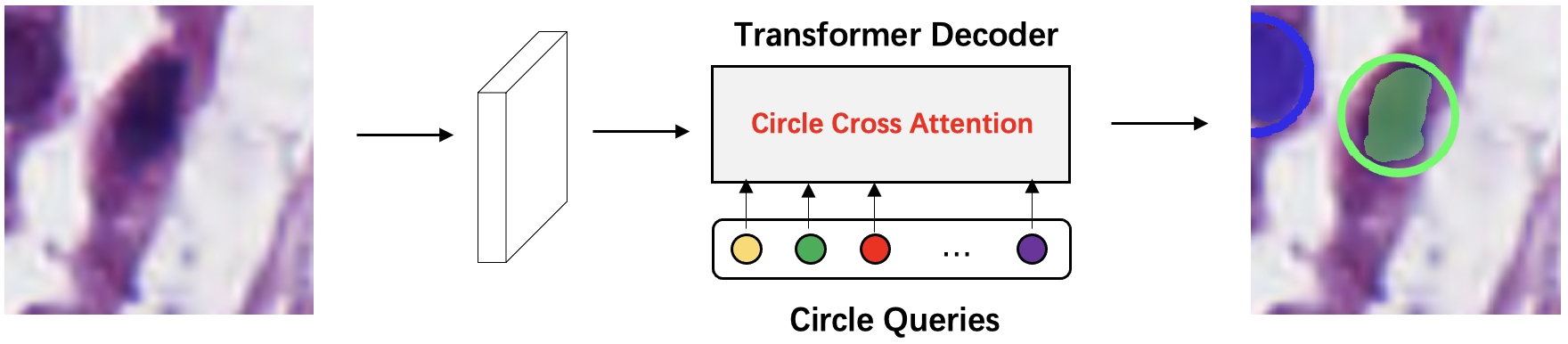}
        		\captionsetup{font={tiny}}
        		\caption{Transformer-based circle detection \& segmentation.}
        		\label{Transformer-based_circle}
			\end{center}
			\end{minipage}
			\setcounter{figure}{2}
			\begin{minipage}[b]{1\linewidth}
			\begin{center}
        		\begin{subfigure}[b]{0.30\linewidth}
        		\includegraphics[width=\linewidth]{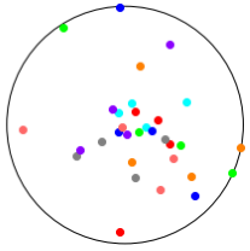}
				\captionsetup{font={tiny}}
        		\caption{CDA-r}
        		\label{CDA-r}
    		\end{subfigure}
    		\begin{subfigure}[b]{0.30\linewidth}
        		\includegraphics[width=\linewidth]{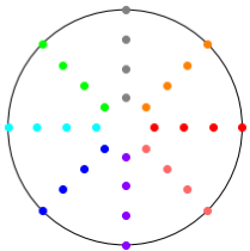}
				\captionsetup{font={tiny}}
        		\caption{CDA-c}
        		\label{CDA-c}
    		\end{subfigure}
			\captionsetup{font={tiny}}
			\caption{initialization of cross-attention module.}
        	\label{multi_head_init}
			\end{center}
			\end{minipage}
			\setcounter{figure}{3}
			\begin{minipage}[b]{1\linewidth}
			\begin{center}
        		\begin{subfigure}[b]{0.30\linewidth}
        		\includegraphics[width=\linewidth]{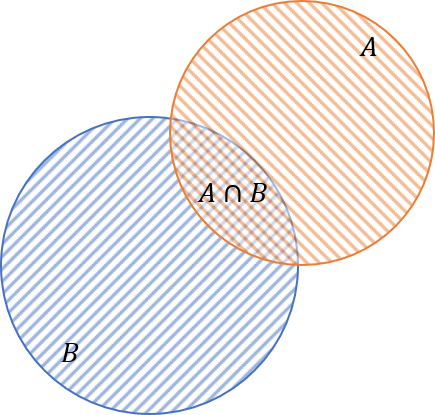}
				\captionsetup{font={tiny}}
        		\caption{CIoU}
        		\label{CIoU}
    		\end{subfigure}
    		\begin{subfigure}[b]{0.30\linewidth}
        		\includegraphics[width=\linewidth]{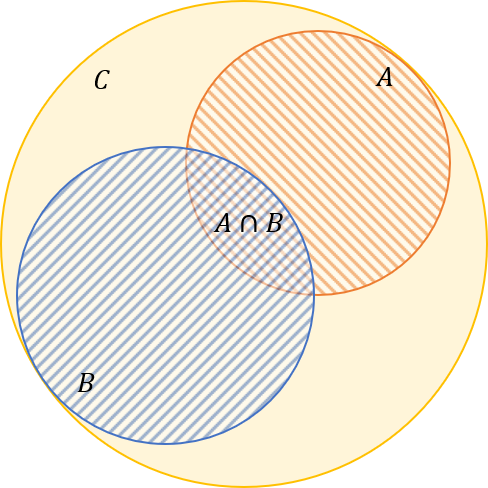}
				\captionsetup{font={tiny}}
        		\caption{gCIoU}
        		\label{gCIoU}
    		\end{subfigure}
			\captionsetup{font={tiny}}
			\caption{Different IoU evaluation metrics.}
        	\label{IoU}
			\end{center}
			\end{minipage}
    \end{minipage}
	\setcounter{figure}{1}
	\begin{minipage}[b]{0.53\linewidth}
			\begin{center}
			\includegraphics[width=\linewidth]{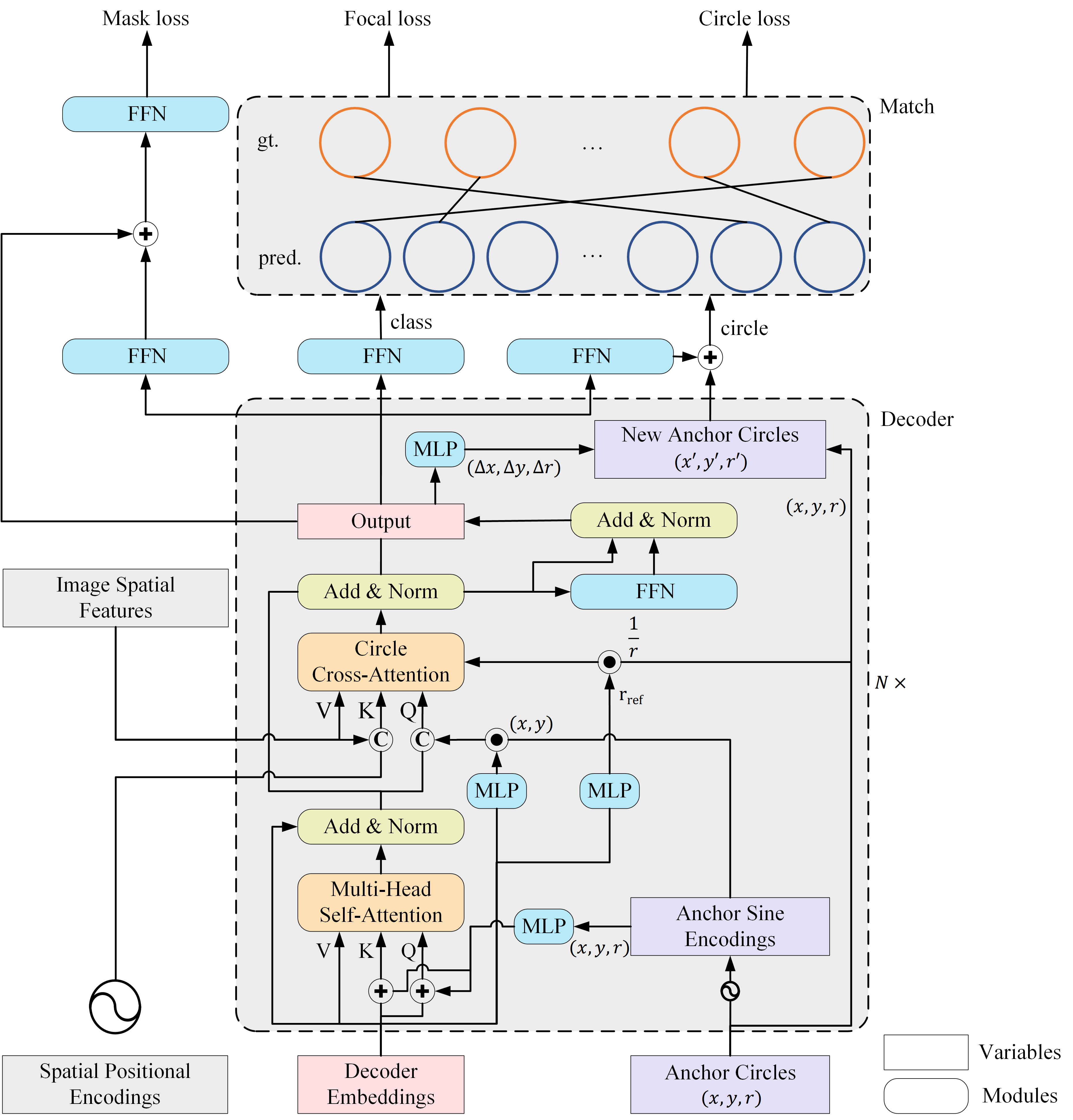} 
			\captionsetup{font={tiny}}
			\caption{Overview of the proposed method.}
			\label{fig_overview}
			\end{center}
    \end{minipage}
 \end{center}
 \end{figure}

Nuclei detection is a highly challenging task and plays an important role in many biological applications such as cancer diagnosis and drug discovery. Rectangle object detection approaches that use CNN have made great progress in the last decade~\cite{girshick2014rich, ren2015faster, tian2019fcos,zhou2019objects,law2018cornernet}. These popular CNN models use boxes to represent objects that are not optimized for circular medical objects, such as detection of glomeruli in renal pathology. To address the problem, an anchor-free CNN-based circular object detection method CircleNet~\cite{yang2020circlenet} is proposed for glomeruli detection. Different from CenterNet~\cite{zhou2019objects},  CircleNet estimates the radius rather than the box size for circular objects. But it also suffers poor detection accuracy for overlapping objects and requires additional post-processing steps to obtain the final detection results.

Recently, DETR~\cite{carion2020end}, a Transformer-based object detection method reformulates object detection as a set-to-set prediction problem, and it removes both the hand-crafted anchors and the non-maximum suppression (NMS) post-processing. Its variants (~\cite{wang2022anchor,2021Deformable,gao2021fast,meng2021conditional,liu2022dabdetr}) demonstrate promising results compared with CNN-based methods and DETR by improving the design of queries for faster training convergence. 
Built upon Conditional-DETR, DAB-DETR~\cite{liu2022dabdetr} introduces an analytic study of how query design affects rectangle object detection. 
Specifically, it models object query as 4D dynamic anchor boxes $(x,y,w,h)$ and iteratively refine them by a sequence of Transformer decoders. 
However, recent studies on Transformer-based detection methods are designed for rectangle object detection in computer vision, which are not specifically designed for circular objects in medical images.

In this paper, we introduce CircleFormer, a Transformer-based circular object detection for medical image analysis. Inspired by DAB-DETR, we propose to use an anchor circle $(x,y,r)$ as  the query for circular object detection, where $(x,y)$ is the center of the circle and $r$ is the radius. We propose a novel circle cross attention module which enables us to apply circle center $(x,y)$ to extract image features around a circle and make use of circle radius to modulate the cross attention map. In addition, a circle matching loss is adopted in the set-to-set prediction part to process circular predictions. In this way, our design of CircleFormer lends itself to circular object detection. We evaluate our CircleFormer on the public MoNuSeg dataset for nuclei detection in whole slide images. Experimental results show that our method outperforms both CNN-based methods for box detection and circular object detection. It also achieves superior results compared with recently Transformer-based box detection approaches. Meanwhile, we carry out ablation studies to demonstrate the effectiveness of each proposed component. To further study the generalization ability of our approach, we add a simple segmentation branch to CircleFormer following the recent query based instance segmentation models\cite{yu2022soit,fang2021instances} and verify its performance on MoNuSeg as well.

\section{Method}

\subsection{Overview}

Our CircleFormer (Fig.~\ref{Transformer-based_circle}) consists of a CNN backbone, a Transformer encoder module, a Transformer decoder and a prediction head to generate circular object results. The detail of the Transformer decoder is illustrated in Fig.~\ref{fig_overview}.

\subsection{Representing Query with Anchor Circle}

Inspired by DAB-DETR, we represent queries in Transformer-based circular object detection with anchor circles. We denote $C_i = (x_i, y_i, r_i)$ as the $i$-th anchor, $x_i, y_i, r_i \in \mathbb{R}$. Its corresponding content part and positional part are $Z_i \in \mathbb{R}^D$ and $P_i \in \mathbb{R}^D$, respectively. The positional query $P_i$ is calculated by: 
\begin{equation}
	P_i = \text{MLP}(\text{PE}(C_i)), \text{PE}(C_i) = \text{PE}(x_i, y_i, r_i) = \text{Concat}(\text{PE}(x_i), \text{PE}(y_i), \text{PE}(r_i)), 
\end{equation}
where positional encoding (PE) generates embeddings from floating point numbers, and the parameters of the MLP are shared among all layers. 

In Transformer decoder, the self-attention and cross-attention are written as:
\begin{equation}
	\text{Self-Attn}: Q_i = Z_i + P_i, K_i = Z_i + P_i, V_i = Z_i
	\label{self attention}
\end{equation}
\begin{equation}
    \begin{split}
    \text{Cross-attn}: Q_i = \text{Concat}(Z_i,\text{PE}(x_i, y_i)\cdot \text{MLP}^{(\text{csq})}(Z_i)) \\
    K_{x,y} = \text{Concat}(F_{x,y}, \text{PE}(x,y)),\ V_{x,y} = F_{x.y},
    \end{split}
    \label{eqn_cross_attn}
\end{equation}
where $F_{x,y} \in \mathbb{R}^D$ denote the image feature at position $(x,y)$ and an $\text{MLP}^\text{(csq)}$ : $\mathbb{R}^D\rightarrow\mathbb{R}^D$ is used to obtain a scaled vector conditioned on content information for a query.

By representing a circle query as $(x,y,r)$, we can refine the circle query layer-by-layer in the Transformer decoder. Specifically, each Transformer decoder estimates relative circle information $(\Delta x, \Delta y, \Delta r)$. In this way, the circle query representation is suitable for circular object detection and is able to accelerate the learning convergence via layer-by-layer refinement scheme.

\subsection{Circle Cross attention}

We propose circle-modulated attention and deformable circle cross attention to consider size information of circular object detection in cross attention module.

\subsubsection{Circle-modulated Attention.} 
The circle radius modulated positional attention map provides benefits to extract image features of objects with different scales. 

\begin{equation}
	\text{MA}((x,y),(x_{ref},y_{ref})) = (\text{PE}(x)\cdot\text{PE}(x_{ref})\frac{r_{i,ref}}{r_i} 
	+\text{PE}(y)\cdot\text{PE}(y_{ref})\frac{r_{i,ref}}{r_i})/\sqrt{D},
	\label{circle-modulated attention}
\end{equation}
where $r_i$ is the radius of the circle anchor $A_i$, and $r_{i,ref}$ is the reference radius calculated by $r_{i,ref} = \text{sigmoid}(\text{MLP}(C_i))$. $\text{sigmoid}$ is used to normalize the prediction $r_{i,ref}$ to the range $[0,1]$. 

\subsubsection{Deformable Circle Cross Attention.}

We modify standard deformable attention to deformable circle cross attention by applying radius information as constraint. Given an input feature map $F \in \mathbb{R}^{C \times H \times W}$, let $i$ index a query element with content feature $Z_i$ and a reference point $P_i$, the deformable circle cross attention feature is calculated by:
\begin{align}
CDA(Z_i,P_i,F) 
	= \sum^M_{m=1} W_m \sum^K_{k=1}Attn_{mik} \dot W'_mF\left((P_{ix} 
	+\Delta r_{mik} \notag \right.\\ \left. 
	\times r_{i,ref} \times 
	\cos{\Delta\theta}_{mik}, P_{iy} 
	+\Delta r_{mik} \times r_{i,ref} \times \sin{\Delta\theta}_{mik})\right),
	\label{Circle Deformable attention}
\end{align}
where $m$ indexes the attention head, $k$ indexes the sampled keys. $M$ and $K$ are the number of multi-heads and the total sampled key number. $W_m' \in \mathbb{R}^{D\times d}$, $W_m \in \mathbb{R}^{d \times D}$ are the learnable weights and $d = D/M$.  $Attn_{mik}$ denotes attention weight of the $k^{th}$ sampling point in the $m^{th}$ attention head. $\Delta r_{mik}$ and $\Delta\theta_{mik}$ are radius offset and angle offset, $r_{i,ref}$ is the reference radius. In circle deformable attention, we transform the offset in polar coordinates to Cartesian coordinates so that the reference point ends up in the circle anchor.

Rather than initialize the reference points by uniformly sampling within the rectangle as does Deformable DETR, we explore two ways to initialize the reference points within a circle, random sampling (CDA-r) and uniform sampling (CDA-c) (As in Fig.~\ref{multi_head_init}). Experiments show that CDA-c initialization of reference points outperforms others.

\subsection{Circle Regression}

A circle is predicted from a decoder embedding as $\hat{c}_i = \text{sigmoid}(\text{FFN}(f_i) + [A_i])$,
where $f$ is the decoder embedding. $\hat{c}_i = (\hat{x}, \ \hat{y}, \ \hat{r})$ consists of the circle center and circle radius. $\text{sigmoid}$ is used to normalize the prediction $\hat{c}$ to the range $[0,1]$. $\text{FFN}$ aims to predict the unnormalized box, $A_i$ is a circle anchor.

\subsection{Circle Instance Segmentation}

A mask is predicted from a decoder embedding by 
$\hat{m}_i = \text{FFN}(\text{FFN}(f_i) + f_i)$,
where $f$ is the decoder embedding. $\hat{m}_i \in \mathbb{R}^{28\times 28}$ is the predicted mask. We use dice and BCE as the segmentation loss: $ \mathcal{L}_{seg} = \lambda_{dice}\mathcal{L}_{dice}(m_i, \hat{m}_i) + \lambda_{bce}\mathcal{L}_{bce}(m_i, \hat{m}_i)$ between prediction $\hat{m}_i$ and the groundtruth $m_i$. 

\subsection{Generalized Circle IoU}
CircleNet extends intersection over union (IoU) of bounding boxes to circle IoU (cIoU) and shows that the cIOU is a valid overlap metric for detection of circular objects in medical images. To address the difficulty optimizing non-overlapping bounding boxes, generalized IoU (GIoU)~\cite{rezatofighi2019generalized} is introduced  as a loss for rectangle object detection tasks. We propose a generalized circle IoU (gCIoU) to compute the similarity between two circles:
$gCIoU = \frac{C_{A}\cap C_{B}}{C_{A}\cup C_{B}} - \frac{|C_{C}-(C_{A}\cup C_{B})|}{C_{C}}$,
where $C_A$ and $C_B$ denotes two circles, and $C_C$ is the smallest circle containing these two circles. 
We show that gCIoU can bring consistent improvement on circular object detection.
Fig.~\ref{IoU} shows the different measurements between two rectangles and circles. Different from CircleNet that only uses cIoU in the evaluation,  we incorporate gCIoU in the training step.
Then, we define the circle loss as: $\mathcal{L}_{circle}(c,\hat{c}) = \lambda_{gciou}\mathcal{L}_{gciou}(c,\hat{c}) 
 	+ \lambda_{c} \| c -\hat{c} \|_1$, 
while $\mathcal{L}_{gciou}$ is generalized circle IoU loss, $\|\cdot\|_1$ is $\ell_1$ loss, and $\lambda_{gciou}, \lambda_{c} \in \mathbb{R}$ are hyperparameters. 

\subsubsection{Circle Training Loss.}
Following DETR, $i$-th each element of the groundtruth set is $y_i = (l_i,c_i)$, where $l_i$ is the target class label (which may be $\varnothing$) and $c_i=(x,y,r)$. We define the matching cost between the predictions and the groundtruth set as:
 \begin{equation}
	\mathcal{L}_{match}(y_i,\hat{y}_{\sigma(i)}) =  \mathbb{I}_{\{l_i\ne\varnothing\}}\lambda_{focal}\mathcal{L}_{focal}(l_i, \hat{l}_{\sigma(i)}) 
	+ \mathbb{I}_{\{l_i\ne\varnothing\}}\mathcal{L}_{circle}(c_i,\hat{c}_{\sigma(i)}),
	\label{match cost}
\end{equation}
where $\sigma \in \mathfrak{S}_N$ is a permutation of all prediction elements, $\hat{y}_{\sigma(i)} = (\hat{l}_{\sigma(i)},\hat{c}_{\sigma(i)})$ is the prediction, $\lambda_{focal}\in \mathbb{R}$ are hyperparameters, and $\mathcal{L}_{focal}$ is focal loss ~\cite{lin2017focal}.

Finally, the overall loss is:
 \begin{equation}
	\mathcal{L}_{loss}(y_i,\hat{y}_{\hat{\sigma}(i)}) = \lambda_{focal}\mathcal{L}_{focal}(l_i, \hat{l}_{\hat{\sigma}(i)}) 
	+ \mathbb{I}_{\{l_i\ne\varnothing\}}\mathcal{L}_{circle}(c_i,\hat{c}_{\hat\sigma(i)}) + \mathcal{L}_{seg},
	\label{loss}
\end{equation}
where $\hat{\sigma}(i)$ is the index of prediction $\hat{y}$ corresponding to the $i$-th ground truth $y$ after completing the match. $m_i$ is the ground truth obtained by RoI Align~\cite{he2017mask} corresponding to $\hat{m}_i$. 

\section{Experiment}

\subsection{Dataset and Evaluation}

\noindent\textbf{MoNuSeg Dataset.} MoNuSeg dataset is a public dataset from the 2018 Multi-Organ Nuclei Segmentation Challenge~\cite{kumar2019multi}. 
It contains 30 training/validataion tissue images sampled from a separate whole slide image of H\&E stained tissue and 14 testing images of lung and brain tissue images. 
Following ~\cite{yang2020circlenet}, we randomly sample 10 patches with size 512 $\times$512 from each image and create 200 training images, 100 validation images and 140 testing images.

\noindent\textbf{Evaluation Metrics.} 
We use $AP$ for nuclei detection evaluation metrics as in in CircleNet~\cite{yang2020circlenet}, and $AP^{m}$ for the instance segmentation evaluation metrics. $S$ and $M$ are used to measure the performance of small scale with area less than $32^2$ and median scale with area between $32^2$ and $96^2$.

\subsection{Implementation Details}
Two variants of our proposed method for nuclei detection, CircleFormer and CircleFormer-D are built with a circle cross attention module and a deformable circle cross attention module, respectively. CircleFormer-D-Joint (Ours) extends CircleFormer-D to include instance segmentation as additional output. All the models are with ResNet50 as backbone and the number of Transformer encoders and decoders is set to 6. The MLPs of the prediction heads share the same parameters.
Since the maximum number of objects per image in the dataset is close to 1000, we set the number of queries to 1000.
The parameter of focal loss for classification is set to $\alpha = 0.25$, $\gamma = 0.1$.
$\lambda_{focal}$ is set to 2.0 in the matching step and $\lambda_{focal} = 1.0$ in the final circle loss.
We use $\lambda_{iou} = 2.0$ , $\lambda_{c}=5.0$, $\lambda_{dice} = 8.0$ and $\lambda_{bce} = 2.0$ in the experiments. All the models are initialized with the COCO pre-trained model~\cite{lin2014microsoft}.

\subsection{Main Results}

\begin{table*}[htb]
	\centering
	\begin{scriptsize}
		\begin{tabular}{l c c c c c c c c}
			\toprule
			Methods & Output & MS &  Backbone & ~~~~~AP$\uparrow$~~~~~ & $AP_{(50)}\uparrow$ & $AP_{(75)}\uparrow$ & $AP_{(S)}\uparrow$ & $AP_{(M)}\uparrow$\\
			\hline
			
			Faster-RCNN~\cite{ren2015faster} & Box & & ResNet-50 & 41.6 & 75.0 & 42.1 & 41.6 & \hlb{\underline{38.3}}\\
			Faster-RCNN~\cite{ren2015faster} & Box & & ResNet-101 & 40.9 & 77.5 & 37.2 & 41.0 & 33.9\\
			CornerNet~\cite{law2018cornernet} & Box & & Hourglass-104 & 24.4 & 52.3 & 18.1 & 32.8 & 6.4\\
			CenterNet-HG~\cite{zhou2019objects} & Box & & Hourglass-104 & 44.7 & 84.6 & 42.7 & 45.1 & \hlr{\textbf{39.5}} \\
			CenterNet-DLA~\cite{zhou2019objects} & Box & & DLA & 39.9 & 82.6 & 31.5 & 40.3 & 33.8\\
			CircleNet-HG~\cite{yang2020circlenet} & Circle & & Hourglass-104 & 48.7 & 85.6 & 50.9 & 49.9 & 33.7 \\
			CircleNet-DLA~\cite{yang2020circlenet} & Circle & & DLA & 48.6 & 85.5 & 51.6 & 49.9 & 30.5 \\
			\hline
			DETR~\cite{carion2020end} & Box & & ResNet50 & 22.6 & 52.3 & 14.6 & 23.9 & 18.2\\
			Deformable-DETR~\cite{2021Deformable} & Box & \checkmark & ResNet50 & 39.5 & 81.0 & 32.7 & 40.2 & 18.5\\
			DAB-DETR~\cite{liu2022dabdetr} & Box & & ResNet50 & 45.7  & 88.9 & 41.9  & 46.3  & 34.8 \\
			DAB-D-DETR~\cite{liu2022dabdetr} & Box & \checkmark & ResNet50 & 49.6 & \textbf{89.5} & 51.5  & 50.1 & 31.9\\
			\hline
			\rowcolor[HTML]{EFEFEF}
			CircleFormer & Circle &  & ResNet50 & 49.7 & 88.8 & 50.9 & 51.1 & 35.4\\
			\rowcolor[HTML]{D5D6D8}
			CircleFormer-D & Circle  & \checkmark & ResNet50 & \hlb{\underline{52.9}} & \hlb{\underline{89.6}} & \hlb{\underline{58.7}} & \hlr{\textbf{54.1}} & 31.7\\
                \hline 			
                \rowcolor[HTML]{D5D6D8}
			CircleFormer-D-Joint & Circle  & \checkmark & ResNet50 & \hlr{\textbf{53.0}} & \hlr{\textbf{90.0}} & \hlr{\textbf{59.0}} & \hlb{\underline{53.9}} & 32.8\\
			
			\bottomrule
		\end{tabular}
\caption{Results of nuclei detection on Monuseg Dataset. Best and second-best results are colored \hlr{\textbf{red}} and \hlb{\underline{blue}}, respectively. MS: Multi-Scale.}
\label{tab_cnn}
	\end{scriptsize}
\end{table*}

\begin{table*}[htb]
	\centering
	\begin{scriptsize}
		\begin{tabular}{l |c c c c c |c c c c c }
			\toprule
                \multirow{2}{*}{} & \multicolumn{5}{c|}{Detection} & \multicolumn{5}{c}{Segmentation} \\ 
			Methods   & $AP$ & $AP_{(50)}$ & $AP_{(75)}$ & $AP_{(S)}$ & $AP_{(M)}$
                          & $AP^{m}$ & $AP_{(50)}^{m}$ & $AP_{(75)}^{m}$ & $AP_{(S)}^{m}$ & $AP_{(M)}^{m}$ \\
			\hline
			QueryInst~\cite{fang2021instances}  & 40.2 & 77.7 & 36.7 & 40.8 & 21.4 & 38.0 & 76.2 & 33.6 & 38.0 & 39.4 \\
			SOIT~\cite{yu2022soit}       & 44.8 & 82.6 & 45.2 & 45.5 & 27.5 & 41.3 & 80.6 & 38.8 & 41.3 & 40.7 \\
			\hline
			Deformable-DETR-Joint  & 45.7 & 86.7 & 43.6 & 46.2 & 28.2 & 43.5 & \textbf{84.8} & 40.5 & 43.5 & 42.3 \\
                \rowcolor[HTML]{D5D6D8}
                CircleFormer-D-Joint & \textbf{53.0} &\textbf{90.0} & \textbf{59.0} & \textbf{53.8}&\textbf{32.8} & \textbf{44.4} & 84.5 & \textbf{43.5} & \textbf{44.4} & \textbf{45.3} \\
			\bottomrule
		\end{tabular}
\caption{Results of nuclei joint detection and segmentation on Monuseg Dataset. All the methods are with ResNet50 as backbone.}
\label{tab_joint_det_seg}
	\end{scriptsize}
\end{table*}

In Table~\ref{tab_cnn}, for nuclei detection, we compare our CircleFormer with CNN-based box detection, CNN-based circle detection and Transformer-based box detection. 
Compared to Faster-RCNN~\cite{ren2015faster} with ResNet50 as backbone, CircleFormer and CircleFormer-D significantly improve box AP by 8.1\% and 11.3\%, respectively.
CircleFormer and CircleFormer-D also surpass CircleNet~\cite{yang2020circlenet} by 1.0\% and 4.2\% box AP. 
In summary, our CircleFormer designed for circular object detection achieves superior performance compared to both CNN-based box detection and CNN-based circle detection approaches.

Our CircleFormer with detection head also yields better performance than Transformer-based methods. DETR can not produce satisfied results due to its low convergence. CircleFormer imporoves box AP by 4.0\% compared to DAB-DETR, and CircleFormer-D improves box AP by 13.4\% and 3.3\% compared to Deformable-DETR and DAB-Deformable-DETR. CircleFormer-D-Joint which jointly outputs detection and segmentation results additionally boosts the detection results of CircleFormer-D.

Experiments of joint nuclei detection and segmentation are listed in Table~\ref{tab_joint_det_seg}. Our method outperforms QueryInst~\cite{fang2021instances}, a CNN-based instance segmentation method and SOIT~\cite{yu2022soit}, an Transformer-based instance segmentation approach. We extend Transformer-based box detection method to provide additional segmentation output inside the detection region, denoted as Deformable-DETR-Joint. Our method with circular query representation largely improves both detection and segmentation results.

To summarize, our method with only detection head outperforms both CNN-based methods and Transformer based approaches in most evaluation metrics for circular nuclei  detection task. Our CircleFormer-D-Joint provides superior results compared to CNN-based and Transformer-based instance segmentation methods. Also, our method with joint detection and segmentation outputs also improves the detection-only setting. We have provided additional visual analysis in the open source code repository.

\begin{table*}[htb]
	\begin{center}
        \captionsetup{font=scriptsize}
		\begin{tiny}  
			\begin{tabular}{l| c c c | c c c c c | c c c c c}
				\toprule
				\multirow{2}{*}{Method} & \multicolumn{3}{c|}{IoU} & \multicolumn{5}{c|}{Cross Attention} & \multicolumn{5}{c}{AP}\\ 
				& gIoU & CIoU & gCIoU & $wh$-MA & $c$-MA & SDA & CDA-r & CDA-c & AP $\uparrow$ & $AP_{(50)}$ $\uparrow$ & $AP_{(75)}$ $\uparrow$ & $AP_{(S)}$ $\uparrow$ & $AP_{(M)}$ $\uparrow$\\
				\midrule
				\multirow{3}*{} 
				& \checkmark & & & \checkmark & & - & - & - & 45.7 & 88.9 & 41.9 & 46.3 & 34.8 \\
				& & \checkmark & & & \checkmark & - & - & - & 48.6 & 88.3 & 48.8 & 49.6 & 30.7 \\
				\rowcolor[HTML]{EFEFEF}
				$\dagger$ & &  & \checkmark & & \checkmark & - & - & - & 49.7 & 88.8 & 50.9 & 51.1 & 35.4 \\
				
				\hline
				\multirow{6}*{} 
				& \checkmark &  & & - & - & \checkmark &  &     & 49.6 & 89.5 & 51.5 & 50.1 & 31.9\\
				& & \checkmark &   & - & - & \checkmark &  &   & 50.8 & 88.2 & 54.4 & 51.7 & 26.6 \\
				& & \checkmark &  & - & - & &\checkmark  &   & 51.1 & 87.9 & 55.8 & 52.5 & 23.2 \\
				& & \checkmark & & - & - & & & \checkmark & 51.1 & 86.8 & 56.6 & 52.7 & 30.1 \\
				& &  & \checkmark & - & - & \checkmark &  &   & 51.1 & 87.6 & 55.6 & 52.7 & 29.1 \\
				& &  & \checkmark & - & - &  & \checkmark &   & 51.8 & 88.4 & 57.2 & 53.1 & 30.2 \\
				\rowcolor[HTML]{D5D6D8}
				$\ddagger$ & & & \checkmark & - & - & & & \checkmark & \textbf{52.9} & \textbf{89.6} & \textbf{58.7} & \textbf{54.1} & 31.7\\
				\bottomrule
			\end{tabular}
		
		\caption{Results of the ablation study analyzing the effects of proposed components in CircleFormer on Monuseg Dataset. $wh$-MA: wh-Modulated Attention; $c$-MA: circle-Modulated Attention; SDA: standard deformable attention; CDA-r: Circle Deformable Attention with random initialization; CDA-c: Circle Deformable Attention with cirle initialization. $\dagger$ denotes CircleFormer. $\ddagger$ denotes CircleFormer-D.}
  \label{tab_components}
       \end{tiny}		
	\end{center}
\end{table*}

\subsection{Ablation Studies}

We conduct ablation studies with CircleFormer on the nuclei detection task.

\begin{table}
\parbox{.45\linewidth}{
\centering
\captionsetup{font=scriptsize}
		\begin{tiny}  
\begin{tabular}{c c c c c c c}
			\toprule
			\# &  AP$\uparrow$ & $AP_{(50)}\uparrow$ & $AP_{(75)}\uparrow$ & $AP_{(S)}\uparrow$ & $AP_{(M)}\uparrow$\\
			\hline
			1 & 50.3 & 86.9 & 54.4 & 51.6 & 30.3\\
			2 & 50.4 & 87.5 & 53.9 & 51.6 & \textbf{33.3}\\
			4 & 51.3 & 88.6 & 54.9 & 52.2 & 31.1\\
			8 & \textbf{52.9} & \textbf{89.6} & \textbf{58.7} & \textbf{54.1} & 
			31.7\\
			16 & 50.4 & 88.2 & 53.6 & 51.6 & 28.3\\
			
			\bottomrule
		\end{tabular}
  \end{tiny}
\caption{Ablation Study of number of Multi-Head.}
\label{tab_multi_head}
}
\hfill
\parbox{.45\linewidth}{
\centering
\captionsetup{font=scriptsize}
\begin{tiny} 
		\begin{tabular}{c c c c c c c}
			\toprule
			\# &  AP$\uparrow$ & $AP_{(50)}\uparrow$ & $AP_{(75)}\uparrow$ & $AP_{(S)}\uparrow$ & $AP_{(M)}\uparrow$\\
			\hline
			1 & 49.3 & 86.4 & 52.5 & 50.6 & 25.6\\
			2 & 51.2 & 88.4 & 55.7 & 52.4 & 29.6\\
			4 & \textbf{52.9} & \textbf{89.6} & \textbf{58.7} & \textbf{54.1} & 
			\textbf{31.7}\\
			8 & 50.8 & 88.1 & 54.7 & 52.0 & 29.2\\
			
			\bottomrule
		\end{tabular}
  \end{tiny}
\caption{Ablation Study of number of reference points.}
\label{tab_reference}
}
\end{table}

\subsubsection{Effects of the Proposed Components.}
For simplicity, we denote the two parts of Table~\ref{tab_components} as P1 and P2.

In CircleFormer, the proposed circle-Modulated attention (c-MA) improves the performance of box AP from 45.7\%  to 48.6\% box AP (Row 1 and Row 2 in P1). 
We replaced circle IoU (CIoU) loss with generalized circle IoU (gCIoU) loss, the performance is further boosted by 2.2\% (Row 2 and Row 3 in P1).

We obtain similar observations of CircleFormer-D. 
When using standard deformable attention (SDA), learning cIoU loss 
gives a 1.2\% improvement on box AP compared to using box IoU (Row 1 and Row 2 in P2). 
Replacing CIoU with gCIoU, the performances of SDA (Row 2 and Row 5 in P2), CDA-r (Row 3 and Row 6 in P2) and CDA-c (Row 4 and Row 7 in P2) are boostd by  0.3\% box AP, 0.7\% box AP and 1.8\% box AP, respectively. Results show that the proposed gCIoU is a favorable loss for circular object detection.

Two multi-head initialization methods, random sampling (CDA-r) and uniform sampling (CDA-c), achieve similar results (Row 3 and Row 4 in P2) and both surpass SDA by 0.3\% box AP (Row 2 and Row 3 in P2).
By using gCIoU, CDA-r and CDA-c initialization methods surpasses SDA 1.1\%  box AP (Row 5 and Row 6 in P2), and 1.8\% box AP (Row 5 and Row 7 in P2), respectively.

\subsubsection{Numbers of Multi-Head \& Reference Points.}
We discuss how the number of Multi-heads in the Decoder affects the CircleFormer-D-DETR model. 
We vary the number of heads for multi-head attention and the performance of the model is shown in the Table ~\ref{tab_multi_head}. We find that the performance increases gradually as the number of heads increases up to 8. However, the performance drops when the number of head is 16. We assume increasing the number of heads brings too many parameters and makes the model difficult to converge. Similarly, we study the impact of the number of reference points in the cross attention module. We find that 4 reference points give the best performance. Therefore, we choose to use 8 attention heads of decoder and use 4 reference points in the cross attention module through all the experiments.

\section{Conclusion}
In this paper, we introduce CircleFormer, a Transformer-based circular medical object detection method. It formulates object queries as anchor circles and refines them layer-by-layer in Transformer decoders. In addition, we also present a circle cross attention module to compute the key-to-image similarity which can not only pool image features at the circle center but also leverage scale information of a circle object. 
We also extend CircleFormer to achieve instance segmentation with circle detection results.
To this end, our CircleFormer is specifically designed for circular object analysis with DETR scheme. 

\section{Acknowledgments}
This work is supported in part by NSFC (61973294), Anhui Provincial Key R$\&$D Program (2022i01020020), and the University Synergy Innovation Program of Anhui Province, China (GXXT-2021-030).

%
%
\bibliographystyle{splncs04}
\bibliography{paper478}
%




\end{document}